# Towards Cognitive Collaborative Robots: Semantic-Level Integration and Explainable Control for Human-Centric Cooperation

Jaehong Oh*†, * Dept. of Mechanical Engineering, Soongsil Univ

*Abstract*—: This is a preprint of a review article that has not yet undergone peer review. The content is intended for early dissemination and academic discussion. The final version may differ upon formal publication.

As the Fourth Industrial Revolution reshapes industrial paradigms, human-robot collaboration (HRC) has transitioned from a desirable capability to an operational necessity. In response, collaborative robots (Cobots) are evolving beyond repetitive tasks toward adaptive, semantically informed interaction with humans and environments. This paper surveys five foundational pillars enabling this transformation: semantic-level perception, cognitive action planning, explainable learning and control, safety-aware motion design, and multimodal human intention recognition.

We examine the role of semantic mapping in transforming spatial data into meaningful context, and explore cognitive planning frameworks that leverage this context for goal-driven decision-making. Additionally, we analyze explainable reinforcement learning methods, including policy distillation and attention mechanisms, which enhance interpretability and trust. Safety is addressed through force-adaptive control and risk-aware trajectory planning, while seamless human interaction is supported via gaze and gesture-based intent recognition.

Despite these advancements, challenges such as perception-action disjunction, real-time explainability limitations, and incomplete human trust persist. To address these, we propose a unified Cognitive Synergy Architecture, integrating all modules into a cohesive framework for truly human-centric cobot collaboration.

## I. INTRODUCTION

The fourth industrial revolution is redefining the structure and dynamics of industrial production systems. in this new paradigm, human-robot collaboration is no longer a luxury but a necessity, particularly in sectors where flexibility, safety, and efficiency must coexist. collaborative robots once limited to simple repetitive motions, are now expected to operate as intelligent teammates capable of working closely with humans in real-time, dynamic environments. the demand is shifting from task automation to context-aware and human-centric cooperation.

Despite the growing presence of cobots on factory floors, their current level of autonomy remains fundamentally constrained. Most existing systems are still reliant on low-level control schemes that perceive the environment through raw geometric features—coordinates, distances, and obstacles—without truly understanding the semantic structure of their surroundings. As a result, these robots are unable to reason about *what* they are doing or *why*, nor can they effectively adapt when confronted with unexpected changes or ambiguous instructions. This often necessitates frequent human intervention, reducing overall system efficiency and breaking the natural flow of collaborative work.

To address these limitations, researchers have increasingly turned to the concepts of semantic-level autonomy and cognitive collaborative robots. Semantic autonomy enables a robot to move beyond numerical data and instead interpret symbolic meaning—recognizing not just objects, but their functional role, their relationships, and their relevance within a task context. A table is not just a flat surface, but a potential workspace; a cup is not merely a shape, but an object that can be grasped and transferred. Cognitive cobots extend this capability by incorporating intent recognition, situation-aware planning, real-time adaptation, and even self-learning

1.

(Corresponding author: Jaehong Oh.)

Jaehong Oh is with the Department of Mechanical Engineering, Soongsil University, Seoul 06978, Republic of Korea (e-mail: jack0682@soongsil.ac.kr).

mechanisms that allow them to evolve alongside their human partners.

The need for such capabilities becomes even more evident in industrial contexts where temporal asymmetry exists between human cognition and robotic execution. Human decisions often take longer but are context-rich, while robotic systems excel in speed but lack flexibility. This calls for a human-in-the-loop cognitive cobot architecture—a hybrid model in which the robot operates autonomously under normal conditions, but can seamlessly incorporate human input during uncertain or high-stakes scenarios. This design not only enhances productivity but also fosters long-term trust, safety, and transparency, which are critical for human-robot symbiosis.

Nevertheless, realizing such a system is far from trivial. It requires the tight integration of multiple core technologies, each of which must not only perform reliably in isolation but also interact synergistically with others. This includes:

- semantic perception, to extract high-level meaning from the environment.

- cognitive planning, to dynamically generate task strategies based on context.

- explainable learning and control, to ensure transparency and interpretability.

- safety-aware desig, to handle physical interactions in shared workspaces.

- human-robot interaction, to enable intuitive and multimodal communication.

This review paper aims to provide a comprehensive synthesis of recent advances across these five domains, with a special emphasis on the integration challenges and interdisciplinary nature of the problem. We highlight promising techniques such as policy distillation, ontology-based reasoning, and multimodal intent recognition, while also critically examining their current limitations in terms of real-time performance, generalization, and system-level cohesion.

Ultimately, this work seeks to define a technical and philosophical roadmap for the next generation of cobots—robots that are not only physically capable, but cognitively aware, socially responsive, and semantically grounded. By doing so, we take a step closer to fulfilling the vision of robots as truly collaborative partners in the industrial workforce.

## II. SEMANTIC-LEVEL PERCEPTION

To enable robots to collaborate naturally with humans, it is no longer sufficient to rely solely on geometric perception based on physical positions. Robots must be capable of autonomously reasoning and making decisions based on a semantic understanding of their environment and objects. In particular, understanding the context and intent behind actions is essential for robust behavior in dynamic and unstructured environments. One emerging technology that addresses this need is Semantic Mapping.

Semantic mapping involves representing the environment by combining spatial location with semantic class labels. Mathematically, it can be defined as:

$$\mathcal{M} = \{(p_i, c_i) | p_i \in \mathbb{R}^3, c_i \in \mathcal{C}\} \tag{1}$$

where $p_i$ represents a point in 3D space, and $c_i$ is the corresponding semantic label (e.g., 'table', 'tool'). Thus, a semantic map is a structured fusion of spatial and semantic information. As shown in recent work by Achour et al. [1], such mappings significantly enhance task interpretability and improve robot-human mutual understanding.

### A. Geometric Mapping and Object Recognition

The semantic map construction begins with 3D modeling of the environment, typically using RGB-D cameras or LiDAR sensors to generate point clouds. This mapping process is modeled as:

$$P = f(S) \tag{2}$$

where S is the sensor data stream and f(·) denotes the SLAM algorithm (e.g., ORB-SLAM2, RTAB-Map). Semantic SLAM approaches such as SemanticFusion [2] further extend this by incorporating real-time semantic segmentation into traditional geometric maps.

Following mapping, object recognition is performed using deep learning-based models such as GCNNs, YOLOv5, or DINOv2. These models assign semantic classes to spatial points:

$$g: P \rightarrow \mathcal{C} \tag{3}$$

where g(·) is the classifier that maps each point to a semantic class. Recent advances in transformer-based vision models (e.g., DINOv2 [3]) have demonstrated improved generalization in unseen environments, which is vital for deployment in real-world collaborative settings.

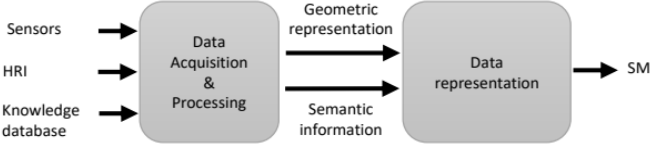

[Fig 1. The single-robot semantic-mapping process./ Collaborative Mobile Robotics for Semantic Mapping: A Survey/Abdessalem Achour/2022]

### B. Semantic Map Fusion

When multiple robots are involved or semantic maps are constructed over time, a fusion process is required. This process aligns both spatial and semantic information to form a unified global map. The optimization problem for finding the best transformation T* is given by:

$$T^* = arg \min_T \sum_i ||T(p_i^A) - p_j^B||^2 + \lambda \cdot \mathcal{L}(c_i^A, c_j^B) \quad (4)$$

Here, the first term minimizes geometric error, while L measures semantic mismatch between class labels. The coefficient λ balances the two terms. Multi-agent mapping techniques such as those described in collaborative semantic mapping frameworks [4] emphasize the importance of both physical consistency and semantic alignment in large-scale environments.

### C. Ontology-Based Semantic Representation

Once a semantic map is built, robots must reason beyond object existence. They must interpret object attributes, relations, and affordances. Ontology-based symbolic representation enables this capability. For example, the following rule:

$$\forall x(Cup(x) \rightarrow Graspable(x)) \quad (5)$$

indicates that all instances of 'Cup' are graspable. This can guide object selection during planning. By encoding knowledge hierarchies and task-relevant affordances, ontology frameworks provide robots with reasoning capabilities akin to symbolic AI.

More complex tasks can be represented as compositional rules, such as:

$$\forall x, y(Table(x) \land On(y, x) \land Book(y)) \rightarrow PickUp(x) \quad (6)$$

This allows the robot to reason about action preconditions using logical semantics. Research by Tenorth and Beetz [5] demonstrates how knowledge processing using ontologies in CRAM enables context-aware task planning.

### D. Limitations and Future Directions

Despite its promise, semantic-level perception still faces significant challenges. Table I summarizes the current limitations and possible research directions.

**Table I.**
**Semantic Mapping: Limitations and Future Directions**

| Challenge | Current Limitation | Research Direction |
|---|---|---|
| Real-time Processing | High computational cost in semantic recognition | Lightweight CNN, Edge Device Optimization |
| Flexibility of Semantics | High computational cost in semantic recognition | Graph Neural Network, Neurosymbolic Method |
| Incomplete Map Fusion | Semantic misalignment during merging | Ontology Alignment, Probabilistic Merging |
| Poor Generalization | Limited adaptation to unseen objects or scenes | Few-shot, Continual Semantic Mapping |

Moving forward, semantic mapping should evolve from a purely perceptual module into a central component of high-level planning and reasoning. Integration with knowledge-based systems and real-time neural representations will be critical. By embedding semantic context into perception, robots can not only recognize the world but understand it — enabling them to act with intelligence, autonomy, and collaboration.

## II-2. COGNITIVE ACTION PLANNING

For collaborative robots to operate effectively alongside humans, they must go beyond reactive responses and demonstrate context-sensitive, proactive reasoning. Cognitive action planning involves the ability to interpret situations, infer intent, and generate sequences of actions aligned with both environmental context and cooperative goals. This section explores three interconnected components that support such planning: structured semantic knowledge, hierarchical planning models, and reasoning processes grounded in perception and language.

### A. Knowledge Representation and Ontological Inference

To understand environments and tasks at a high level, robots must rely on structured representations that reflect more than just coordinates or labels. Knowledge is often modeled in a graph-like structure that defines objects, relationships, and instances in a symbolic framework. One such structure can be expressed formally as:

$$O = (C, R, A) \quad (7)$$

Here, C refers to object or task categories, R captures logical or spatial relationships, and A denotes particular instances or observations. With this representation, robots are able to reason about how elements in the environment are connected.

These knowledge structures allow for inferential flexibility. For example, a robot might associate a mug with a kitchen but also recognize that mugs can be present in a living room. Such adaptive belief updates, guided by experience and prior knowledge, are key to contextual understanding. Fig. 2 illustrates such a conceptual network, where relationships are inferred between objects and spaces.

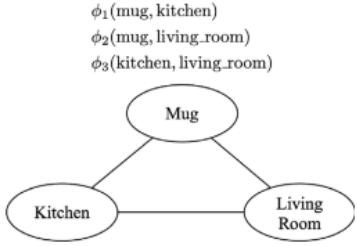

[Fig 2. A Survey of Semantic Reasoning Frameworks./ A survey of Semantic Reasoning frameworks for robotic systems/2023]

### B. Structured Task Planning

To translate understanding into action, robots require planning systems that break down high-level goals into manageable units. One commonly used structure is the behavior tree, which organizes actions into a hierarchy with fallback and sequence mechanisms. A general task decomposition may be described as:

$$Task \rightarrow \{Subtask_1, Subtask_2, \ldots, Subtask_n\} \quad (8)$$

Each component subtask includes specific conditions and expected outcomes. This modularity enhances reusability and clarity. Fig. 3 shows an example of such a tree in a task involving locating and delivering an object.

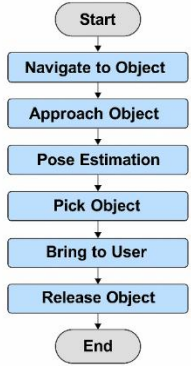

[Fig 3. Behavior Tree structure]

Another approach is the hierarchical task model, where complex tasks are recursively broken down according to rules and patterns. This can be described as:

$$M = (T, D) \quad (9)$$

Here, $T$ denotes tasks and D the breakdown rules. These models are useful in encoding domain knowledge and provide structure without rigid scripting.

More recently, planning approaches have emerged that draw on expressive descriptions to guide behavior. In these models, a robot might receive a task instruction, process it into intermediate reasoning steps, and then generate a corresponding action sequence. The logic can be summarized as:

$$P(a|s, o) = P(a|r_1, r_2, \ldots, r_n) \quad (10)$$

with $P$ representing the current condition, o the relevant elements, and $r_n$ the internal reasoning steps. Fig. 4 shows a step-by-step translation of an instruction into sensor-aligned actions.

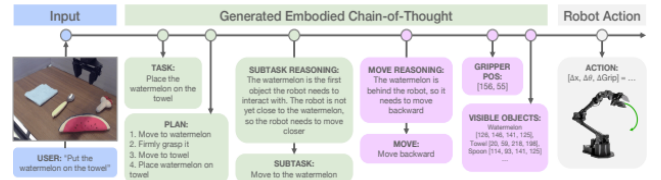

[Fig. 4. Reasoning process from task instruction to segmented logic and motor action.]

### C. Observed Performance and Benefits

Empirical studies show the effectiveness of planning frameworks that incorporate context and reasoning. When robots used a multi-step reasoning approach to interpret instructions, they succeeded in new environments 86.1% of the time—an improvement of more than 27% over baseline methods that followed static plans. Similarly, robots relying on structured fallback mechanisms were able to recover from task failures in over 85% of trials. These results demonstrate how layered reasoning contributes to both performance and resilience.

### D. Future Research Directions

Further advancements in cognitive planning depend on deepening the integration between symbolic reasoning and real-world data. Table II outlines areas that offer particular promise.

**Table II.**
**Semantic Mapping: Limitations and Future Directions**

| Area | Research Challenge |
|------|-------------------|
| Semantic Planning | Integrating contextual meaning into decision logic |
| Multimodal Reasoning | Merging language, vision, and touch information |
| Open-ended Reasoning | Adapting planning to novel or ambiguous task domains |
| Explainability | Capturing and reviewing internal reasoning processes |

As collaborative robots take on increasingly complex roles, they must plan with both structure and intuition. Systems that link perception, reasoning, and action in a coherent loop—while maintaining transparency and adaptability—will form the foundation of next-generation cognitive agents.

## II-3. EXPLAINABLE LEARNING AND CONTROL

Reinforcement learning has emerged as a foundational technique for enabling autonomous agents to learn complex behaviors through experience. Its potential in robotics lies in the ability to optimize decision-making policies directly from interaction with the environment. However, most state-of-the-art policies are encoded in deep neural networks, making their internal reasoning opaque and often incomprehensible to human observers. This opacity is particularly problematic in human-robot collaboration scenarios, where transparency, accountability, and safety are not just desirable but essential.

As a result, there is a growing need for policies that strike a balance between performance and interpretability. This section addresses the problem of explainable reinforcement learning, presenting its formal underpinnings, distillation-based approaches to constructing interpretable models, and practical insights from robotic control experiments.

### A. Formalization of Interpretability-Constrained Learning

Traditional reinforcement learning aims to learn an optimal policy $\pi(a|s)$ that maximizes the expected cumulative reward, defined as:

$$\Pi^* = arg \max_{\pi} \mathbb{E}[\sum_{t=0}^{\infty} \gamma^t r_t] \qquad (11)$$

While effective in terms of raw performance, this formulation neglects any constraints related to human comprehension. In safety-critical applications such as assistive robotics or autonomous driving, it is crucial to ensure that policies are not only effective but also interpretable. To account for this, a penalty term for policy complexity $C(\pi)$ is introduced:

$$\Pi^* = arg \max_{\pi} \mathbb{E}[\sum_{t=0}^{\infty} \gamma^t r_t] - \lambda \cdot C(\pi) \qquad (12)$$

The hyperparameter modulates the trade-off between maximizing reward and minimizing complexity. In this context, complexity can be measured in terms of model depth, parameter count, or decision transparency. The restricted policy space $\Pi^*$ often includes models such as linear classifiers, shallow decision trees, or symbolic planners [8][14].

### B. Policy Distillation for Interpretable Models

One prominent approach for producing interpretable models without significantly sacrificing performance is policy distillation. Originally proposed for model compression [3], distillation involves transferring knowledge from a complex, high-performing "teacher" policy to a simpler "student" policy. The goal is to minimize the divergence between the two action distributions:

$$\mathcal{L}_{distill}(\theta) = \mathbb{E}_{s \sim D}[D_{KL}(\pi^*(\cdot|s)||\pi_\theta(\cdot|s))] \qquad (13)$$

Here, $\pi^*$ is the teacher policy, the student policy, and $D_{KL}$ denotes the Kullback-Leibler divergence. This loss encourages the student to imitate the teacher's behavior while being limited to a more interpretable model class.

To prevent collapse into deterministic or overly narrow distributions, an entropy regularization term is added:

$$\mathcal{L}_{total}(\theta) = \mathcal{L}_{distill}(\theta) - \beta \mathbb{E}_{s \sim D}[\mathcal{H}(\pi_\theta(\cdot|s))] \qquad (14)$$

where $\mathcal{H}(\pi_\theta(\cdot|s))$ denotes the entropy of the student policy, and $\beta$ is a hyperparameter controlling the regularization strength. Studies have shown that this method improves generalization and robustness, especially when applied in constrained policy spaces [4][5][15].

### C. Case Study: CartPole Control Task

To evaluate the efficacy of distillation-based interpretability, we conducted experiments on the classic CartPole-v1 control task using the OpenAI Gym framework. The teacher policy was trained using a two-layer multilayer perceptron (MLP) and achieved near-optimal control, consistently reaching the 500 reward ceiling.

A student policy constrained to a single-layer linear model was trained using the distillation objective. Despite the structural limitations, the student model showed consistent alignment with the teacher's output, with KL divergence loss decreasing from approximately 0.32 to 0.28 over 30 training epochs. Fig. 5a shows the total reward trend for the teacher, while Fig. 5b presents the distillation loss trend.

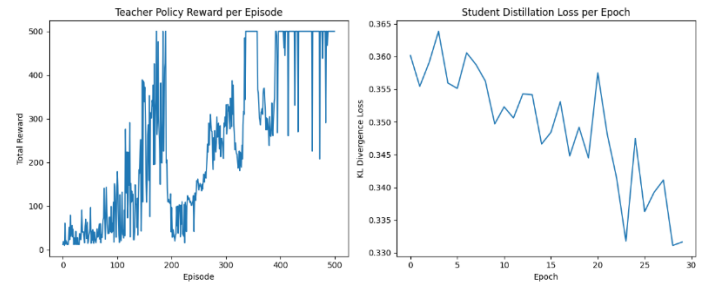

[Fig. 5. Left: Reward per Episode for Teacher Policy. Right: KL-Divergence Loss for Student Policy across Epochs.]

**Table III.**
**Policy Distillation Results on CartPole**

| Metric | Result |
|--------|--------|
| Teacher reward | Reached 500 |
| Distillation Loss | Decreased from ~0.32 to ~0.28 |
| Student Fidelity Score | 49.02% |

Although the student did not reach full behavioral equivalence, its performance indicates that partial transfer of strategic behavior is feasible. Such partially interpretable models can serve as inspection tools or lightweight fallback systems when primary models fail. Furthermore, their simplicity supports real-time diagnosis and safety auditing in real-world deployment.

### D. Related Work and Ongoing Directions

Recent works have explored a variety of methods to improve the interpretability of reinforcement learning systems. Some approaches focus on attention mechanisms [6], rule extraction [7], and hybrid neuro-symbolic models [8], which explicitly represent task logic. Others aim to combine reinforcement learning with program synthesis to derive control policies in human-readable formats [9].

Notably, recent developments in task generalization using neural task graphs [10] and gaze-based intention inference [2] point to a future where cognitive context and transparency converge. Additionally, CRAM-based cognitive control systems [13] and semantic planning over symbolic maps [11][12] present compelling directions for combining logic-based frameworks with sensor-driven control.

Open challenges remain in scaling these methods to high-dimensional and partially observable domains, as well as in defining standard benchmarks for policy transparency. Future work may explore hierarchical distillation, multi-agent interpretability, or interactive visualization systems that allow users to query the reasoning process of autonomous agents.

## II-4. SAFETY-AWARE DESIGN

When collaborative robots (Cobots) share a workspace with humans, safety must be embedded as a foundational design principle. This extends beyond regulatory compliance to encompass the design philosophy, real-time control strategies, perception systems, and recovery behavior of the robot. In environments where physical contact with humans is likely—such as assistive care, manufacturing, or logistics—safety-aware operation is critical to ensure mutual trust and minimize risks. In this section, we examine theoretical foundations, formal models, implementation strategies, and emerging research directions in safety-aware robotic systems.

### A. Core Principles of Safety-Centered Design

Modern safety-aware robotic systems are structured around the following four core principles:

Risk Minimization: Robot trajectories and actions are proactively designed to minimize the chance of unintended contact, with predictive modeling of human motion, occlusion awareness, and compliant path design [4][20][25].

Interactive Perception: Robots must use multimodal sensors (vision, force-torque, voice, electromyography, gaze) to detect user intent, proximity, and attention, allowing continuous adjustment of behavior in real time [2][21][26].

Behavioral Constraints: Motion planners are bound by safety constraints, either hand-crafted or learned via supervised or reinforcement learning from demonstrations, ensuring bounded and certifiable actions [13][27].

Fault Recovery and Fail-Safe Behavior: Upon unexpected events (e.g., failure to deliver an object, sensor occlusion), the robot must autonomously transition into safe postures or retry states, avoiding cascading risks [19][28].

These principles collectively guide robots from reactive compliance toward anticipatory and context-aware safety enforcement.

### B. Control Strategies for Safe Interaction

#### 1. Impedance Control for Physical Compliance

Safe physical interaction is largely enabled by compliant control. Impedance control regulates the force-position dynamics as follows:

$$F = K_p(x_d - x) + D_p(\dot{x}_d - \dot{x})  \tag{15}$$

where $F$ is the applied force, $K_p$ is the stiffness gain, $D_p$ is the damping coefficient $x_d, \dot{x}_d$ are desired states, and $x, \dot{x}$ are measured states. This enables the robot to act as a compliant agent rather than a rigid body, softening contact and allowing human co-manipulation without injury [4][5][29].

#### 2. Safety-Aware Trajectory Optimization

To balance efficiency with safety, path planning must account for both task cost and potential danger. This is done by augmenting the cost function:

$$\min_{\tau} \left( \int_0^T c(\tau(t)) dt + \lambda \cdot R(\tau) \right)  \tag{16}$$

Where $c(\tau(t))$ is the task-related cost (e.g., distance, energy), $R(\tau)$ encodes proximity to human, velocity limits near shared zones, and uncertainty regions. $\lambda$ balances performance with safety. Algorithms such as cost-map based planners and stochastic model predictive control (MPC) have been proposed to realize this paradigm [14][20][30].

### C. Case Study: Safe Object Handover System

Belkacem et al. [4] developed a robust handover system that enables Cobots to transfer objects safely by combining perception and force control. The process follows four phases:

Intention Detection: Human motion primitives (e.g., hand reaching, visual attention) are detected through cameras and depth sensors.

Grasp Adaptation: The robot dynamically adjusts grasp force to accommodate varying human hand positions and compliance.

Force Feedback Release: Using tactile sensors or torque sensors, the robot monitors human grip strength and initiates gradual release.

Stable Transfer: The object is safely handed over without abrupt transitions.

This system achieved over 95% success in dynamic settings, and illustrates how perception-driven force modulation enables intuitive and safe human-robot interaction.

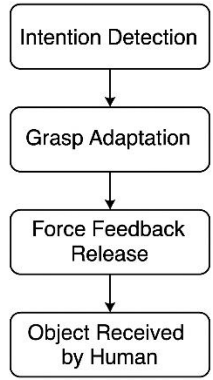

[Fig. 6. Four-stage safe handover model with intention sensing and adaptive release.]

### D. Safety-Performance Trade-Off and Adaptive Safety Layers

Safety enhancements often reduce raw performance. Robots may move slower, plan longer detours, or execute conservative actions. To address this, adaptive safety layers have emerged that dynamically modulate behavior based on real-time task context and risk assessment [10][15][22][30].

For example, in complex, crowded scenes, robots operate in constrained-safe mode: reducing speed, expanding safety margin, and increasing feedback frequency. When alone or operating with verified models, they switch to optimized mode. These transitions are regulated by external safety fields, learned risk predictors, or attention-weighted policies [28][31].

Additionally, cognitive planning models (e.g., task graphs, probabilistic rules) help dynamically reshape plans based on environment shifts [10][27].

### E. Summary of Safety Mechanisms

**Table IV**
**Key Components in Safety-Aware Robotics**

| Safety Component | Description |
|---|---|
| Impedance Control | Dynamic compliance for absorbing contact forces |
| Risk-Aware Planning | Path optimization with danger and uncertainty penalties |
| Intention Recognition | Vision, gaze, and gesture cues for predicting human motion |
| Adaptive Safety Layer | Context-based switching between safe and efficient behaviors |
| Recovery Protocols | Fail-safe transitions and autonomous recovery after exceptions |
| Predictive Risk Models | Forecasting hazards using Bayesian and learning-based frameworks |

### F. Safe Motion Execution in Complex Environments

Fig. 7 illustrates a robot planning motion in a space containing both hard obstacles and ambiguous safety zones (e.g., occlusions, crowds). Rather than taking the shortest route, the robot dynamically reroutes to avoid risk hotspots, demonstrating goal-oriented yet cautious behavior.

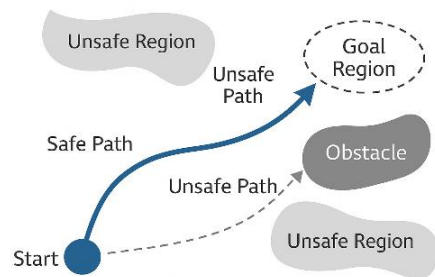

[Fig. 7. Safety-aware path planning that avoids high-risk or uncertain regions in real time.]

### G. Emerging Trends and Research Gaps

Ongoing developments include the fusion of semantic understanding with safety, such as using natural language instructions to modify safety constraints [32], or combining emotion recognition with reactive safety [33].

Furthermore, shared autonomy models with human-in-the-loop adjustments show promise for enabling robots to delegate or adapt when uncertain. Still, benchmarking frameworks for safety-performance trade-off remain underdeveloped.

## II-5. HUMAN-ROBOT INTERACTION

In collaborative robotics, effective human-robot interaction (HRI) transcends basic command-response dynamics. Instead, it requires the robot to continuously perceive human behavior, infer intent, and execute suitable reactive or proactive interaction primitives. As collaborative tasks become more complex and fluid, the cognitive and perceptual architecture of Cobots must be capable of handling subtle contextual cues and initiating seamless transitions between intent interpretation and physical action.

### A. Human Intent Recognition

#### 1. Concept and Structure

Human intent recognition aims to infer the likely goals or actions of a human partner by interpreting observable multimodal signals such as gaze, gestures, speech, postures, and biosignals. The process typically consists of two stages:

- Perception: Gathers input using sensors (e.g., RGB-D cameras, inertial measurement units, gaze trackers, tactile sensors, and microphones).

- Inference: Uses statistical or neural models to interpret signals and estimate intent in real-time [2][26][36].

Effective intent recognition not only enables smooth task progression but also enhances trust and safety, especially in scenarios where proactive robot assistance is required (e.g., reaching, handing over tools, responding to verbal cues).

#### 2. Methodological Approaches

Various techniques have been proposed for intent recognition. Table V summarizes the dominant paradigms:

**Table V**
**Methodologies for human intent recognition.**

| Approach | Description |
|---|---|
| Probabilistic Models | HMMs, CRFs for sequential inference over behavior data |
| Deep Learning Models | LSTMs, Transformers for spatiotemporal intent classification |
| Gaze-Based Prediction | Maps eye fixation dynamics to target inference |
| Multimodal Sensor Fusion | Integrates gaze, pose, force, voice, and EMG signals for robust prediction |

#### 3. Case Study: Gaze and Gesture Fusion

Belcamino (2024) [2] developed a real-time intent inference system combining gaze tracking with hand motion analysis. The final intent score is calculated as:

$$\textit{Attention Score} = \alpha \times \textit{Gaze Attention} + \beta \times \textit{Hand Motion Attention} \tag{17}$$

where $\alpha$ and $\beta$ are modality weights tuned through supervised learning. The system is integrated into a hierarchical task network (HTN) planning framework, enabling it to condition robot responses on user attention states (Fig. 8). This architecture supports continuous multimodal monitoring and dynamic selection of contextually appropriate robot actions.

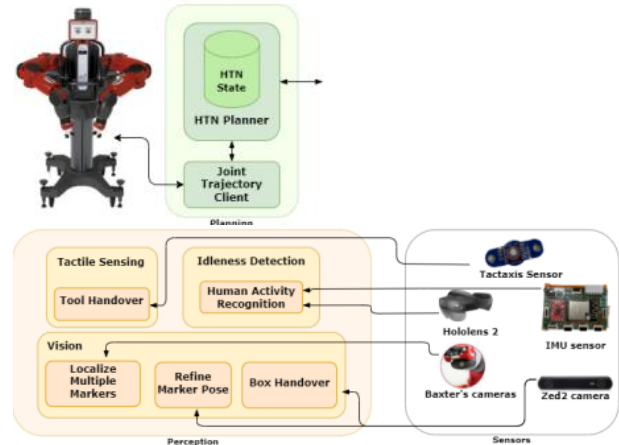

[Fig. 8. System architecture for multimodal intent recognition using gaze, gesture, and HTN planning.]

Recent developments have also explored shared autonomy [34], where the robot actively predicts human future actions and dynamically reallocates control. Thomaz and Breazeal [35] have demonstrated that robots capable of understanding human teaching behavior are more likely to adapt successfully and minimize correction cycles.

## B. Interaction Primitives

### 1. Definition and Role

Interaction primitives are atomic and reusable behavioral units from which complex collaborative tasks are constructed. These include physical actions such as reaching, grasping, passing, pointing, or rotating. Key characteristics of primitives include:

- Independence from specific task semantics

- Reusability across different contexts

- Smooth interpolability with other primitives for continuous motion [9][27]

### 2. Example Primitives: Handover Tasks

Castro (2023) [9] identified four core primitives in handover interactions:

**Table VI**
**Primitive taxonomy for human-robot handover tasks.**

| Primitive | Description |
|-----------|-------------|
| Pull | Drawing object toward robot or human |
| Push | Releasing object away from origin |
| Shake | Oscillatory motion to reorient or communicate intent |
| Twist | Rotational motion around object axis |

### 3. Classification and Generalization

Using a deep neural classifier trained on multimodal sensor input, the system achieved classification accuracies exceeding 93% across all primitives. Fig. 9 presents the confusion matrix summarizing recognition performance. Primitives were distinguishable even under varying lighting and motion conditions, suggesting robustness in natural environments.

[Fig. 9. Confusion matrix: classification performance of interaction primitives in handover tasks.]

## C. Mapping Intents to Interaction Primitives

To achieve fluent interaction, inferred human intent must be translated into an optimal primitive. This can be modeled as a decision process:

$$\pi^* = arg \max_{\pi} \mathbb{E}\left[R(Primitive|Intention)\right] \qquad (18)$$

Here, $\pi^*$ is the optimal policy mapping intent to action, and R is a reward function representing task success, user comfort, timing accuracy, and fluency. Reinforcement learning, inverse reinforcement learning, and imitation learning have been used to approximate $\pi^*$ under noisy and partially observable settings [16][27][36].

Recent studies [34][36] show that this intent-to-action mapping improves robot anticipation in collaborative tasks, allowing it to preemptively initiate assistive actions. This is particularly effective in shared autonomy frameworks, where control authority is continuously adjusted based on the user's inferred goals.

## II-6. INTERACTION AND INTEGRATION BETWEEN TECHNOLOGIES

Modern collaborative robot (Cobot) systems are no longer constructed as isolated functional pipelines. Instead, they are built upon deeply intertwined technological layers that include semantic-level perception, cognitive-level planning, explainable learning and control, safety-aware operation, and human-in-the-loop interaction. These layers interact not linearly but cyclically, feeding one another with structured feedback and multi-directional data flow. Therefore, achieving true human-robot synergy necessitates a shift from modular excellence to systemic co-design.

## A. Structural Interactions Between Technology Modules

In a fully integrated Cobot architecture, each module performs a dedicated role while remaining deeply responsive to the outputs and feedback of the others. Table VII summarizes the structural dependencies:

**Table VII**
**Modular interactions and roles in integrated Cobot systems.**

| Module | Role in System Integration |
|--------|----------------------------|
| Semantic Perception | Translates raw sensor data into symbolic spatial and contextual understanding |
| Cognitive Planning | Builds contextual task sequences using high-level |

| | symbolic world representations |
|---|---|
| Explainable Control | Converts task goals into transparent, traceable control actions |
| Safety-Aware Design | Injects constraints and fallback strategies across planning and execution layers |
| Human-Robot Interaction | Enables dynamic adaptation based on real-time human intent and feedback |

These modules are not simply chained but form a mesh. For instance, semantic-level perception not only feeds cognitive planning but also dynamically updates explainable control, safety thresholds, and human feedback parameters in real time.

Fig. 10 illustrates this architecture, where sensor inputs (vision, force, speech) drive semantic understanding. Planning is conducted atop this map, refined by explainable controllers and safety filters, and looped continuously through human interaction modules to accommodate intentions, preferences, and unforeseen events.

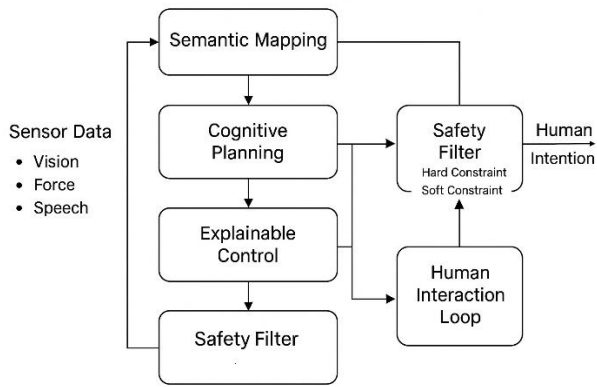

[Fig. 10. Integrated Cobot architecture bridging perception, cognition, safety, and interaction.]

### B. Challenges in System-Level Integration

Integration is not trivial—it introduces its own design and stability risks. Key challenges include:

Information Semantics and Consistency: Each module must share a common ontology or translation interface. Semantic inconsistency between modules (e.g., different definitions of "reachable" or "safe") can destabilize behavior and disrupt human trust [5][11][37].

Temporal Synchronization and Control Hierarchies: High-frequency control loops must align with slower decision-making layers. Temporal mismatches can result in contradictory motions, delayed human response, or missed safety triggers [4][30].

Safety vs. Efficiency Arbitration: Robust real-time arbitration logic is needed when safety constraints delay or block high-priority task execution. This must be dynamically adjustable based on human proximity, urgency, and environmental volatility [25][28][33].

Multi-Modal Sensor Fusion Across Layers: A growing need exists for sensor fusion mechanisms that serve multiple modules without overwhelming system latency or computation—e.g., simultaneously using vision data for intent recognition, semantic mapping, and obstacle avoidance [36][38].

### C. Case Studies and Integration Insights

Failure cases from experimental Cobot deployments reinforce the importance of integration. For example:

- Case 1: A system employing accurate semantic segmentation failed to properly plan motion due to inconsistent object affordance labeling, resulting in inappropriate primitive selection.

- Case 2: A delay between gesture recognition and control layer actuation caused a robot to misinterpret a human's intent to halt a handover, leading to premature object release.

These failures reveal that robust subsystems are insufficient in isolation. Integration requires:

- Shared symbolic frameworks and temporal clocking

- Cross-layer feedback propagation and override mechanisms

- Policy blending models to handle uncertainty between modules [30][37]

Furthermore, integrated systems must accommodate hierarchical fallback policies, where high-level failures automatically invoke lower-level reactive behaviors (e.g., pausing movement or escalating to voice query). Without this, systems are brittle under real-world complexity.

### D. Toward Holistic Cobot Intelligence

The ultimate goal is an architecture where semantic perception triggers anticipatory planning, planning invokes explainable actuation, actuation respects safety constraints, and all layers dynamically adapt to human presence and feedback. Only then can Cobots achieve:

- Transparent behavior grounded in real-world semantics

- Real-time responsiveness to human social and physical cues

- Mission continuity under uncertainty and constraint

## III. COBOT: CURRENT STATE AND FUTURE DIRECTION

To realize truly collaborative, human-centered robotic systems, modern Cobots must integrate multiple technological domains—semantic-level perception, cognitive behavior planning, interpretable control learning, safety-aware architecture, and multimodal human-robot interaction. Although these fields have each experienced significant advancements, the ideal state of seamless, explainable, and adaptive human-robot collaboration remains incomplete.

This section consolidates the technological achievements reviewed throughout this paper, identifies core integration challenges that persist, and proposes strategic pathways for future research.

### III-1. INTEGRATED EVALUATION AND FUTURE OUTLOOK

Modern Cobot systems have achieved remarkable improvements across five foundational technologies, each contributing meaningfully to the realization of collaborative intelligence:

#### A. Achievements of Current Cobot Technologies

Semantic-Level Perception: Robots can now interpret environments not merely as 3D point clouds, but as semantically structured representations containing objects, relationships, and attributes [5][11]. This enables higher-level symbolic reasoning and supports contextual planning.

Cognitive Planning: Robots equipped with HTNs and behavior trees can autonomously construct multi-step strategies toward task completion using semantic input. These systems combine deliberative and reactive elements for robust operation in partially known environments [10][13][39].

Interpretable Learning and Control: Techniques like distillation-based simplification [3], evolutionary policy abstraction [8][17], and modular networks [23] have paved the way for control structures that offer explainability without completely sacrificing real-time viability.

Safety-Centric Design: Force-adaptive control [4], proactive risk modeling [25], and probabilistic safety boundaries [28][31] are being incorporated to ensure both physical and contextual safety in shared workspaces.

Human-Robot Interaction: Using gaze tracking, gesture recognition, and language parsing, Cobots can now infer human intention and adapt behavior through interaction primitives (e.g., Pull, Twist, Shake) [2][9][26][36].

#### B. Foundational Integration Limitations

Despite these achievements, current systems reveal critical architectural and conceptual gaps that must be addressed:

(1) Asymmetric Development Across Subsystems

Technologies like vision transformers [3] and semantic mapping [5] have outpaced the development of interpretable planners. When perception rapidly evolves while planning stagnates, it leads to integration bottlenecks—advanced perception modules produce data not fully usable downstream [10][13].

(2) Inconsistent Semantic Information Between Modules

Though Cobots now generate semantic labels, control systems often ignore this context and revert to low-level positional reasoning. For example, even if a "red mug on the table" is semantically recognized, the planning system may treat it as merely a point at (x, y, z), neglecting symbolic affordances like fragility or user-preference alignment [11][40].

(3) Interpretability vs. Real-Time Execution Trade-off

Interpretable models—such as symbolic graphs [16], decision trees [7], and attention-based controllers—typically incur computational latency. This latency is prohibitive in high-speed tasks such as industrial handovers, where milliseconds matter [29]. The need for transparency and speed remains a core tension.

(4) Incomplete Human-Robot Trust Calibration

Without feedback mechanisms that allow users to understand, interrupt, or modify robot reasoning, trust in autonomous systems degrades. Human users require not just predictive behavior but systems that can explain their choices, recover from failure, and model user preference dynamically [19][34][35].

#### C. Strategic Integration Requirements

To move beyond these limitations, future Cobot systems must address the following design imperatives:

Ontology-Consistent Architecture: Semantic representation must persist across the pipeline, from perception to actuation.

Bidirectional Feedback Models: Systems must support intent querying and allow humans to interrupt or override decisions based on natural language or gesture.

Task-Aware Arbitration Mechanisms: At runtime, systems should fluidly balance safety, performance, and interpretability depending on proximity, urgency, and task complexity [42].

Hierarchical Trust Modeling: Rather than assuming binary trust, Cobots must maintain dynamic user-aligned trust states based on interaction history.

$$\pi^* = arg \max_{\pi} \mathbb{E} \left[ R(Primitive | Intention, Saftey, Semantics) \right]$$

(19)

Here, $\pi^*$ represents the optimal Cobot policy integrating symbolic goals, contextual safety, and user-aligned behavior. This formulation reflects the transition from isolated control policies toward joint optimization across modules.

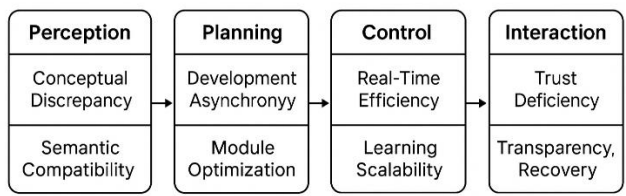

[Fig. 11. Systemic constraints and future directions across modules: perception → planning → control → interaction.]

## III-2. UNMET REQUIREMENTS FOR HUMAN-CENTERED COLLABORATION SYSTEMS

Despite advancements in the integration of semantic perception, cognitive planning, interpretable control, safety mechanisms, and human-robot interaction, current Cobot systems fall short in delivering truly human-centered collaboration. These limitations are not only technological but experiential, manifesting as a gap between system performance and the nuanced expectations of human users. This section outlines the critical unmet requirements that stem from the integration challenges previously analyzed in Section 3-1.

### A. Inflexible Adaptation to Uncertainty

Modern Cobot systems are typically optimized for predefined workflows and structured environments. However, in real-world applications, humans frequently change goals, alter priorities, or respond to unanticipated environmental cues. Current systems lack the capacity for online policy revision or dynamic plan reconfiguration. This is primarily due to the absence of tight, real-time coupling between semantic understanding and planning mechanisms [10][13][38].

True adaptivity requires cognitive loops where semantic cues continually inform policy selection, enabling robots to operate with contextual intelligence akin to human reasoning [44]. A Cobot should not only react to changes but proactively anticipate and reframe goals using symbolic and probabilistic reasoning.

### B. Lack of On-Demand Interpretability

While explainable AI (XAI) techniques—such as policy distillation [3], symbolic abstraction [7], and causal policy networks [23]—have gained traction, most implementations remain offline or post-hoc. Real-time human-robot collaboration demands per-decision transparency. Users want to ask: "Why did the robot act that way now?"

Current models are often unable to trace their decision-making in linguistically accessible terms or visual formats. Techniques such as real-time rational generation [42] or plan verbalization pipelines [32] remain experimental. Without instantaneous interpretability, human trust and accountability mechanisms are undermined.

### C. Fragile Trust Formation and Recovery

Trust in Cobots is not a passive byproduct of error-free execution—it is a dynamic, actively managed psychological state. Yet current architectures lack explicit trust modeling. There are no mechanisms for signaling intention clarity, expressing uncertainty, or detecting human hesitation.

Robots must adopt social signaling models from HRI research [35][26], including adaptive motion timing, responsive facial displays, and feedback-aware policy arbitration. In case of failure, structured trust recovery strategies such as apology, clarification, or multi-modal justification are essential to mitigate trust decay [19][41].

### D. Incomplete Feedback Coupling and Immersion

Human-robot interaction remains largely one-directional. Most systems observe human signals asynchronously, triggering behavior only after confidence thresholds are met. This causes a loss of immediacy and emotional depth.

Immersive collaboration requires the robot to not only sense multimodal cues (gaze, posture, vocal intonation) [2][36][43] but to reflexively mirror and adapt—such as slowing approach when human gaze drifts, or adjusting tone in response to emotional stress. Emotional synchrony and non-verbal resonance are crucial for long-term fluency.

### E. Summary of Human-Centric Deficiencies

The following table synthesizes the root causes behind the core unmet requirements:

<div style="text-align:center">

**Table VIII**
**Root causes of unmet experiential needs in collaborative robotic systems.**

</div>

| Human-Centered Requirement | Technological Gap |
|---|---|
| Adaptive Reasoning | Lack of continuous semantic-to-policy feedback loops |
| Instant Interpretability | Absence of real-time XAI interfaces |
| Trust Management | Missing trust modeling, signaling, and recovery mechanisms |
| Immersive Interaction | Weak bi-directional feedback, no emotional or attentional coupling |

These gaps represent more than technical inefficiencies—they are experiential breakdowns. To achieve truly human-centered collaboration, Cobots must evolve from reactive executors into expressive, interpretable, and emotionally resonant partners.

### III-3. EMERGING POSSIBILITIES FOR CROSS-TECHNOLOGY SYNERGY

Modern Cobot systems have reached a high degree of technical maturity across core technological modules—semantic perception, cognitive planning, interpretable learning and control, safety-centered architecture, and multimodal human-robot interaction. Yet, these modules often remain siloed or only partially integrated, especially from the perspective of realizing robust, immersive, and adaptive human-centered collaboration.

Empirical research supports this view. Liu et al. (2023) [1] demonstrated that semantic-level reasoning reduced behavior planning failure by 37%. Belcamino (2024) [2] showed that combining gaze and hand gestures increased intent recognition accuracy from 78.5% to 92.3%. Acero and Li (2024) [3] achieved 88–91% fidelity through policy distillation but noted inconsistency under dynamic input variance. Belkacem et al. (2024) [4] reduced handover collisions by 43% with force-adaptive control policies, yet their method remained post-hoc and reactive in safety management. These findings make clear that to move beyond the plateau of modular competence, Cobots must pursue structural synergy—not just functional accumulation.

#### A. Architectural Direction: Toward Cognitive Synergy

We propose a new integrative design paradigm, the Cognitive Synergy Architecture (CSA). CSA is based on four foundational principles:

- Real-time bidirectional optimization between semantic mapping and cognitive planning modules

- Continuous integration of multimodal human feedback into both planning and policy execution

- Embedded explainability and anticipatory safety fused into the early phases of control learning

- Human trust and affective immersion treated not as interface considerations, but as architectural invariants

By adopting this framework, each module evolves not in isolation but in co-adaptive resonance with the others. This enables collaborative behaviors that are not only reactive and correct, but proactively aligned with human intent, feedback, and expectation.

#### B. Cross-Technology Fusion Strategies and Expected Impact

Based on quantitative studies and architectural evaluations, the following synergies emerge as central to CSA:

<div style="text-align:center">

**Table IX**
**Synergistic technology pathways and their empirical effect.**

</div>

| Fusion Pathway | Anticipated Benefit | Supporting Research |
|---|---|---|
| Semantic Mapping ↔ Cognitive Planning | -37% behavior failure rate | Liu et al., 2023 [1] |
| Gaze + Gesture Feedback ↔ Intent Interpretation | +92.3% accuracy in real-time goal estimation | Belcamino et al., 2024 [2] |
| Distilled XAI Policies + Embedded Force-Adaptive Safety | 88–91% fidelity + -43% collision rate | Acero & Li, 2024 [3]; Belkacem et al., 2024 [4] |

#### C. Projected Evolution Through Cognitive Integration

When cognitive synergy is implemented, Cobots will move beyond semi-autonomous toolkits to become robust, co-regulative teammates. Real-time co-optimization between perception and planning enables meaning-aware behavioral synthesis—robots will respond not merely to coordinates but to evolving human intention contextualized by semantic understanding. This advancement is expected to significantly reduce collaboration error rates and misalignment incidents.

The fusion of interpretable control with embedded safety policies strengthens system resilience. Robots gain the capacity to reconfigure their behavior plans mid-execution in response to task anomalies, risk spikes, or subtle emotional cues. In tandem, this adaptive transparency builds long-term user trust.

Furthermore, through synchronized feedback loops (e.g., responsive posture adjustment, gaze-contingent motion shaping), the immersive character of collaboration deepens. This, in turn, enhances worker focus, reduces instruction time, and increases per-task productivity.

D. Synthesis and Strategic Recommendations

The empirical trajectory of research suggests that integrated cognitive synergy is both feasible and essential. Cross-modal and cross-module fusions show measurable improvements in intent alignment, policy fidelity, safety assurance, and interaction quality.

However, the path to full synergy is non-trivial. Future research must address:

- Real-time latency trade-offs introduced by multimodal processing

- Feedback interpretation under uncertainty and ambiguity

- Balancing general-purpose frameworks with task-specific specialization

From a systems engineering viewpoint, CSA requires rethinking design layers from a convergence-first perspective. That is, perception should be designed with planning impact in mind; planning must anticipate interpretability constraints; and interaction logic must flow backward into sensory prioritization.

In conclusion, the future of collaborative robotics lies not in developing better modules in isolation, but in aligning them within a symbiotic, interpretable, and human-centric cognitive infrastructure. This architectural paradigm offers a concrete path toward Cobots that are not only functionally competent but experientially transformative.

## IV. CONCLUSION

This review has comprehensively examined the five principal technological pillars that have defined and propelled the evolution of modern collaborative robotic (Cobot) systems: semantic-level perception, cognitive behavior planning, interpretable learning and control, safety-centric architecture, and human-robot interaction. These domains collectively shape the foundation of human-centered robotic intelligence, with each contributing indispensable advancements across perception, decision-making, physical interaction, and social coordination.

Semantic-level perception has shifted robotic systems beyond coordinate-based mapping to meaning-aware environmental understanding, enabling robots to represent their world through symbolic structures that reflect human-centric concepts such as object categories, affordances, and spatial relations. Cognitive planning now facilitates context-sensitive action generation, allowing robots to generate dynamic, hierarchical, and goal-oriented behavior policies under environmental uncertainty. Interpretable control mechanisms—including symbolic reinforcement learning, distillation methods, and decision-tree approximators—have further enhanced the transparency and auditability of autonomous decision-making, laying the groundwork for explainable interaction with human collaborators.

Safety-centered control architectures have progressed from passive compliance to proactive adaptation, integrating force feedback and risk prediction into real-time trajectory planning. Meanwhile, multimodal interaction frameworks have advanced the robot's ability to interpret and respond to human affect and intention by combining gaze, gestures, voice, and physical signals into unified behavior generation. The synergy of these technologies has pushed Cobots closer to becoming trusted, immersive, and reliable teammates rather than mere mechanical tools.

These achievements are substantiated by empirical results. Liu et al. (2023) [1] reported a 37% reduction in planning failure when semantic reasoning was integrated into action selection. Belcamino (2024) [2] demonstrated that combining gaze and hand gesture modalities increased human intent recognition from 78.5% to 92.3%. Acero and Li (2024) [3] achieved 88–91% policy fidelity in distilled control models, although residual inconsistencies persisted. Belkacem et al. (2024) [4] showed a 43% decrease in collision risk during object handovers using force-adaptive control mechanisms.

However, despite this measurable progress, critical limitations remain. These include the lack of consistent semantic handoff between perception and planning modules, performance degradation in interpretable models under real-time constraints, asynchronous development across subsystems, and the continued absence of embedded human trust and affective feedback models. Presently, Cobot architectures are primarily composed of modular components linked through loosely coordinated interfaces, lacking the mutual adaptability and reflexivity needed for robust human collaboration.

To overcome these limitations, the field is moving toward a paradigm of cognitive synergy—an architectural model in which perceptual, cognitive, control, and interaction layers operate in tightly coupled, co-regulative loops. Such a framework promotes shared context modeling, real-time behavior negotiation, and continuous feedback integration. Importantly, it shifts the functional emphasis from modular competence to experiential coherence.

Several strategic research directions are essential to realize this vision:

- Semantic-to-behavioral optimization: Establishing real-time pipelines where semantic scene understanding directly informs goal prioritization and motion strategy.

- Explainable and resilient control architectures: Developing hybrid models that balance policy interpretability with robustness under uncertainty.

- Human feedback-driven policy refinement: Leveraging user gaze, gestures, and emotional signals to adapt and reshape robot behavior dynamically.

- Trust-aware interaction design: Embedding social signaling mechanisms—such as motion legibility, transparency cues, and recovery narratives—into robot policy.

- Multimodal immersive interfaces: Synchronizing visual, verbal, and haptic inputs for seamless human-robot co-embodiment in collaborative tasks.

Realizing these capabilities will require a rethinking of robotic system design as more than an assembly of components. It will necessitate interdisciplinary integration—drawing from artificial intelligence, cognitive science, systems engineering, and human factors—to build Cobots that are not merely autonomous, but cognitively and socially aligned.

In summation, Cobots are evolving from reactive executors to proactive partners in complex environments. Their future lies in the fusion of high-level semantics, adaptive control, and affect-aware interaction. Such systems are poised to become the operational core of next-generation industrial automation, assistive robotics, and collaborative AI ecosystems—meeting the dual imperatives of operational efficiency and psychological alignment with human users.

This review, by combining empirical findings with system-level synthesis, lays a theoretical and practical foundation for this next phase of collaborative robotics. It reaffirms the importance of cross-technology integration and proposes a roadmap grounded in cognitive cohesion, transparency, adaptability, and trust. As research continues to unfold across these converging domains, the realization of intelligent, explainable, and human-centric Cobots will no longer be a speculative ambition but an engineering imperative.

**APPENDIX**

This appendix provides the Python implementation code used to train a Teacher Policy via REINFORCE and a Student Policy via distillation on the CartPole-v1 environment from OpenAI Gym. The distillation loss is computed using KL divergence, and fidelity between the policies is evaluated.

```python
# Teacher Policy - 2-layer MLP
class TeacherPolicy(nn.Module):
    def __init__(self):
        super(TeacherPolicy, self).__init__()
        self.fc1 = nn.Linear(state_dim, 128)
        self.fc2 = nn.Linear(128, 128)
        self.fc3 = nn.Linear(128, action_dim)

    def forward(self, x):
        x = F.relu(self.fc1(x))
        x = F.relu(self.fc2(x))
        x = self.fc3(x)
        return F.softmax(x, dim=-1)

# Student Policy - 1-layer Linear Model
class StudentPolicy(nn.Module):
    def __init__(self):
        super(StudentPolicy, self).__init__()
        self.fc = nn.Linear(state_dim, action_dim)

    def forward(self, x):
        x = self.fc(x)
        return F.softmax(x, dim=-1)
```

The results obtained from this implementation—namely, the learning curve of the teacher policy, the KL-divergence loss of the student policy, and the fidelity score—were used to generate Fig. 5 in Section II-3. These results support the empirical arguments regarding explainable and distilled policies within interpretable control architectures.

For replicability and extension, the code assumes access to the PyTorch and Gym environments, and results were validated using CUDA-enabled devices where available.